\documentclass[aapm,mph,preprint,graphicx]{revtex4-1}

\draft 

\usepackage{graphicx}
\usepackage{gensymb} 
\usepackage{amsfonts, amsmath, amssymb}
\usepackage{booktabs}       
\usepackage{dcolumn}
\usepackage{bm}
\usepackage[utf8]{inputenc}
\usepackage{algorithm}
\usepackage{algcompatible}
\usepackage{algpseudocode}
\usepackage{multirow}
\usepackage{natbib}

\usepackage{array}
\newcolumntype{L}[1]{>{\raggedright\let\newline\\\arraybackslash\hspace{0pt}}m{#1}}
\newcolumntype{C}[1]{>{\centering\let\newline\\\arraybackslash\hspace{0pt}}m{#1}}
\newcolumntype{R}[1]{>{\raggedleft\let\newline\\\arraybackslash\hspace{0pt}}m{#1}}

\algnewcommand\algorithmicto{\textbf{to}}

\newcommand*{\thead}[1]{%
\multicolumn{1}{c}{\begin{tabular}{@{}c@{}}#1\end{tabular}}}

\begin{document}


\title{Deep CT to MR Synthesis using Paired and Unpaired Data} 



\author{Cheng-Bin Jin}
\affiliation{School of Information and Communication Engineering, INHA University, Incheon 22212, South Korea}

\author{Hakil Kim}
\affiliation{School of Information and Communication Engineering, INHA University, Incheon 22212, South Korea}

\author{Wonmo Jung}
\affiliation{Acupuncture and Meridian Science Research Center, Kyung Hee University, Seoul 02447, South Korea}

\author{Seongsu Joo}
\affiliation{Team Elysium Inc., Seoul 93525, South Korea}

\author{Eunsik Park}
\affiliation{Team Elysium Inc., Seoul 93525, South Korea}

\author{Young Saem Ahn}
\affiliation{Department of Computer Engineering, INHA University, Incheon 22212, South Korea}

\author{In Ho Han}
\affiliation{Department of Neurosurgery, Medical Research Institute, Pusan National University Hospital, Pusan 49241, South Korea}

\author{Jae Il Lee}
\affiliation{Department of Neurosurgery, Medical Research Institute, Pusan National University Hospital, Pusan 49241, South Korea}

\author{Xuenan Cui}
\affiliation{School of Information and Communication Engineering, INHA University, Incheon 22212, South Korea}


\date{\today}

\begin{abstract}
\textbf{Purpose:} Computed tomography (CT)-based radiotherapy is currently used in radiotherapy planning and its effect is quite good. Magnetic resonance (MR) imaging will play a very important role in radiotherapy treatment planning for segmentation of tumor volumes and organs. However, the use of MR-based radiotherapy is limited because of the high cost and the increased use of metal implants such as cardiac pacemakers and artificial joints in aging society. In addition, CT scans can also differentiate soft tissue, especially with an intravenous contrast agent, and has higher imaging resolution, and less motion artifact due to its high imaging speed, which are its advantages compared with MR imaging. To improve the accuracy of CT-based radiotherapy planning, we propose a synthetic approach to produce synthesized MR images from brain CT images.

\noindent
\textbf{Methods:} A MR synthetic generative adversarial network (MR-GAN) was trained to transform 2D brain CT image slices into 2D brain MR image slices, combining adversarial loss, dual cycle-consistent loss, and voxel-wise loss. The MR-GAN base on the objective functions has a dual cycle-consistent term for paired and unpaired training data. The dual cycle-consistent term includes four cycles: forward unpaired-data, backward unpaired-data, forward paired-data, and backward paired-data cycles. Both networks in MR-GAN were trained simultaneously with discriminators estimating the probability that a sample came from real data rather than the synthesis networks, while the synthesis networks were trained to translate realistic synthetic data that could not be distinguished from real data by the discriminators.

\noindent
\textbf{Results:} The experiments were analyzed using CT and MR images of 202 patients. Qualitative and quantitative comparisons against independent paired training and unpaired training methods demonstrate the superiority of our approach. Unlike other methods, the proposed approach utilizes the adversarial loss from a discriminator network, dual cycle-consistent loss using paired and unpaired training data, and voxel-wise loss based on paired data to synthesize realistically-looking MR images. Quantitative evaluation results show that the average correspondence between synthesized and reference MR images in our approach is much better than in other methods; synthesized images are closer to the reference, and achieve the lowest MAE of $19.36 \pm 2.73$ and the highest PSNR of $65.35 \pm 0.86$. 

\noindent
\textbf{Conclusions:} Our approach uses paired and unpaired data to solve the context-misalignment problem of unpaired training, and alleviate the rigid registration task and blurred results of paired training. Unpaired data is plentifully available, and together with limited paired data, could be used for effective synthesis in many cases. Our results on the test set demonstrate that MR-GAN was much closer to the reference MR images when compared with other methods. The preliminary results indicated that the synthetic system is able to efficiently translate structures within complicated 2D brain slices, such as soft brain vessels, gyri, and bones. The proposed approach can potentially increase the quality of synthesized images for a synthetic system that depends on supervised and unsupervised settings, and can also be extended to support other applications, such as MR-CT and CT-PET synthesis.

\vspace{0.5cm}
\end{abstract}


\maketitle 

\section{Introduction} \label{Intro}
CT-based radiotherapy is currently used in radiotherapy planning and its effect is quite good. Recently, radiotherapy devices using magnetic resonance (MR) imaging are being developed, since MR imaging is much better than computed tomography (CT) scan in the contrast of soft tissue. In particular, the use of MR-based radiotherapy is increasing in brain tumors, and MR imaging will play a very important role in radiotherapy planning in the near future.  However, MR imaging usually costs more than a CT scan, and the time required for a complete MR scan takes about 20 to 30 minutes. Conversely, a CT scan is usually completed within 5 minutes. In addition, CT scans can also differentiate soft tissue, especially with an intravenous contrast agent, and has higher imaging resolution, and less motion artifact due to its high imaging speed, which are its advantages compared with MR imaging. Furthermore, the use of MR-based radiotherapy has been limited in situations where the use of metal implants such as cardiac pacemakers and artificial joints is increasing in aging society. Much of the concern about CT scans is the harm of radiation. However, there is no risk to patients, even for a patient with lung tuberculosis who undergoes several X-rays in one year. The real risk is to professionals–technicians and radiologists. Of course, this is a controversial topic among experts. In this paper, we propose a synthetic approach to produce synthesized MR images from brain CT images. To the best of our knowledge, this is the first study that attempts to translate a CT image to an MR image.

The major contributions of this paper can be summarized as follows:
\begin{itemize} 
\setlength\itemsep{0em}
\item The proposed approach uses paired and unpaired data to overcome the context-misalignment issue of unpaired training, and to alleviate the registration and blurry results of paired training.
\item The paper introduces MR-GAN framework by combining adversarial loss, dual cycle-consistent loss, and voxel-wise loss for training paired and unpaired data together.
\item The proposed approach can be easily extended to other data synthesis (MR-CT and CT-PET synthesis) to benefit the medical image community.
\end{itemize}

\newpage
Recently, advances in deep learning and machine learning in medical computer-aided diagnosis (CAD) \cite{son2017retinal, chen2017dcan}, have allowed systems to provide information on potential abnormalities in medical images. Many methods have synthesized a CT image from the available MR image for MR-only radiotherapy treatment planning \cite{edmund2017review}. The MR-based synthetic CT generation method \cite{han2017mr} used deep convolutional neural networks (CNN) with paired data, which was rigidly aligned by the minimization of voxel-wise differences between CT and MR images. However, minimizing the voxel-wise loss between the synthesized image and the reference image during training may lead to blurry generated outputs. In order to obtain clear results, Nie et al. \cite{nie2017medical} proposed a method that combined the voxel-wise loss with an adversarial loss in a generative adversarial network (GAN) \cite{goodfellow2014generative}. Concurrent work \cite{bi2017synthesis} proposed a similar idea to synthesize positron emission tomography (PET) images from CT images using multiple channel information of the pix2pix framework by Isola et al. \cite{isola2017image}. Ben-Cohen et al. \cite{ben2017virtual} combined fully convolutional network (FCN) \cite{long2015fully} and the pix2pix \cite{isola2017image} model to export initial results and blend the two outputs to generate a synthesized PET image from a CT image.

\begin{figure}
  \centering
  \includegraphics[width=1\linewidth]{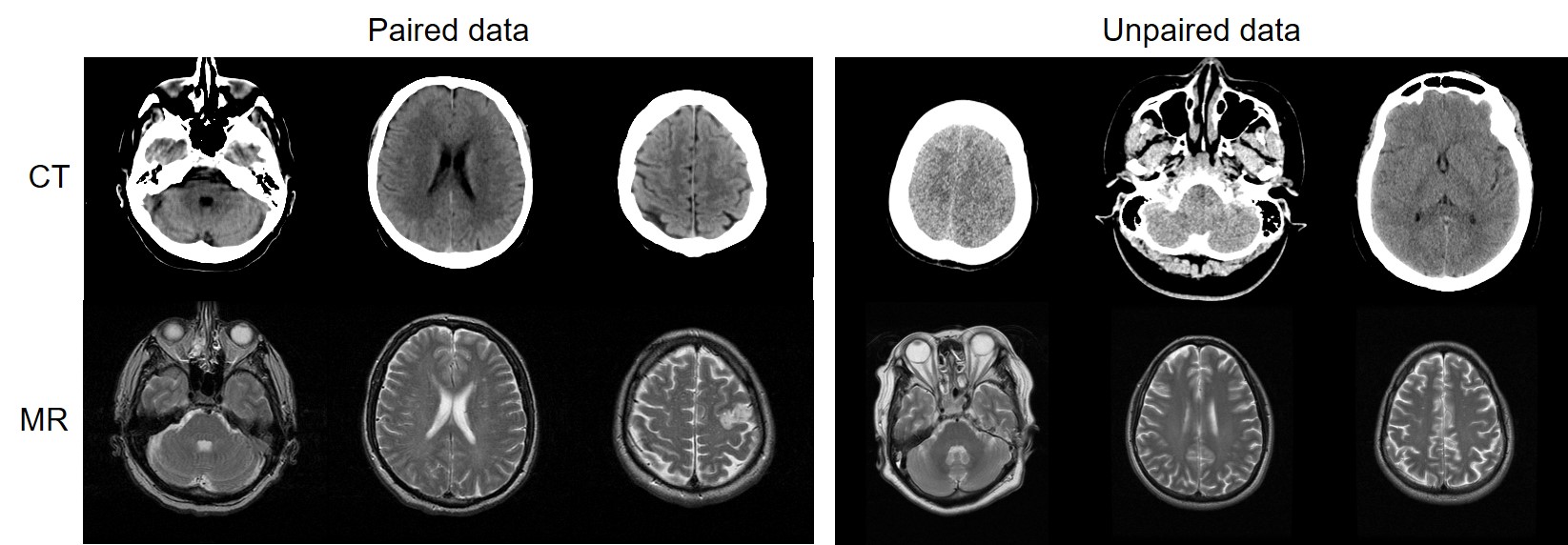}
  \caption{{\it Left} Deep networks train with paired data, which include CT and MR slices taken from the same patient at the same anatomical location. Paired data need to be intentionally collected and aligned, which imposes difficulty. However, paired data give network regression constraints that are far more correct. {\it Right} Deep networks train with unpaired data, which include CT and MR slices that are taken from different patients at different anatomical locations. There is a considerable amount of unpaired data.}
  \label{fig01}
\end{figure}

Although the combination of the voxel-wise loss with adversarial loss addresses the problem of blurry generated synthesis, the voxel-wise loss is dependent on the availability of large numbers of aligned CT and MR images. Obtaining rigidly aligned data can be difficult and expensive. However, most medical institutions have considerable unpaired data that were scanned for different purposes and different radiotherapy treatments. Using unpaired data would increase the amount of training data exponentially, and alleviate many of the constraints of current deep learning-based synthetic systems  (Fig. \ref{fig01}). Unlike the paired data-based methods in \cite{han2017mr, nie2017medical, bi2017synthesis, ben2017virtual}, Wolterink et al. \cite{wolterink2017deep} used a CycleGAN model \cite{zhu2017unpaired}, which is an image-to-image translation with unpaired images used to synthesize CT images from MR images. In an unpaired GAN paradigm, we want the synthesized image to not only look real, but also to be paired up with an input image in a meaningful way. Therefore, cycle-consistency loss is enforced to translate the synthesized image back to the original image domain, and minimize the difference between the input and the reconstructed image as a regularization. Because of the large amount of unpaired data, the synthesized images are more realistic than the results from the paired training methods. However, compared to the voxel-wise loss of the paired data, the cycle-consistent loss still has certain limitations in correctly translating the contextual information of soft tissues and blood vessels.
\section{Materials and Methods} \label{data}
\subsection{Data acquisition}
\begin{figure}
  \centering
  \includegraphics[width=1\linewidth]{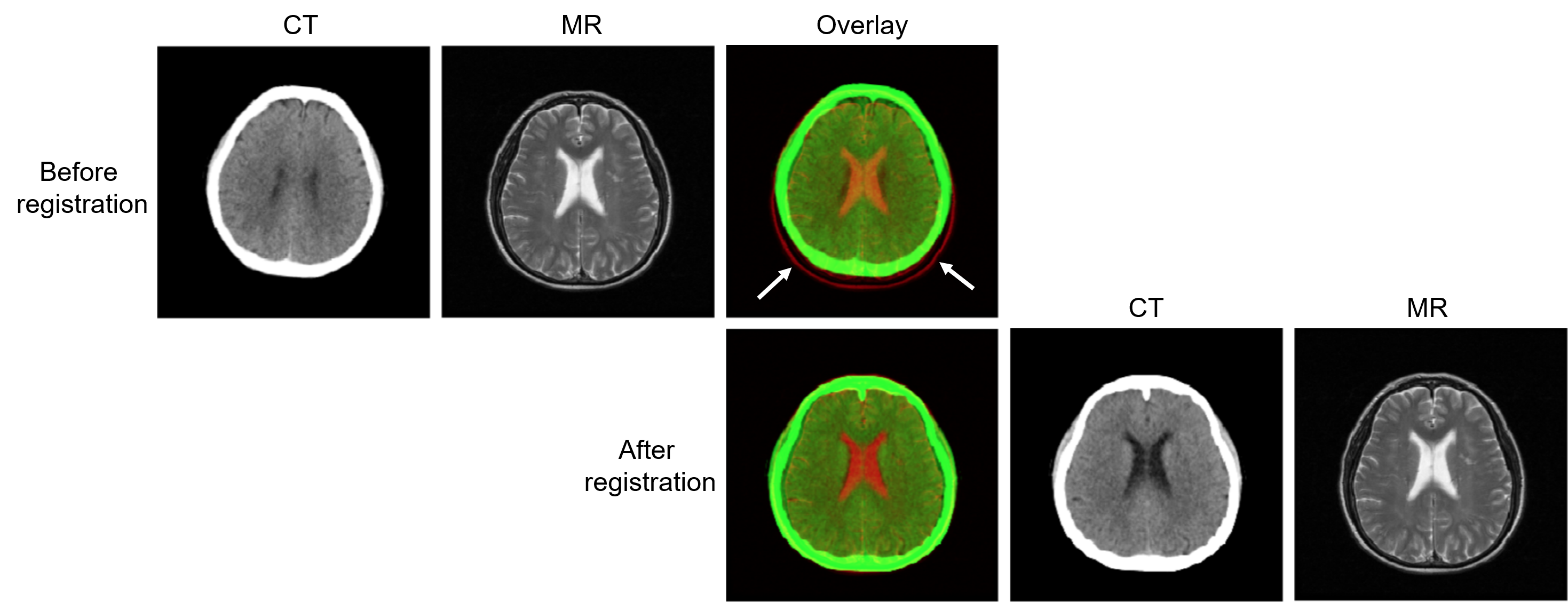}
  \caption{Examples showing registration between CT and MR images after the mutual-information affine transform.}
  \label{fig02}
\end{figure}

Our dataset consisted of the brain CT and MR images of 202 patients who were scanned for radiotherapy treatment planning for brain tumors. Among these patients, 98 patients had only CT images, and 84 patients had only MR images. These data belonged to the unpaired data. For the remaining 20 patients, CT and MR images were acquired during radiation treatment. CT images were acquired helically on a GE Revolution CT scanner (GE Healthcare, Chicago, Illinois, United States) at 120 kVp and 450 mA. T2 3D MR (repetition time, 4320 ms; echo time, 95 ms; flip angle 150\degree) images were obtained with a Siemens 3.0T Trio TIM MR scanner (Siemens, Erlangen, Germany). To generate paired sets of CT and MR images, CT and MR images of the same patient were aligned and registered using affine transformation based on mutual information. CT and MR images were resampled to the same voxel size $\left(1.00 \times 1.00 \times 1.00 \ mm^{3}\right)$. Before the registration, the skull area in the CT images was removed by masking all voxels above a manually selected threshold. Skull-stripped MR brain image were also registered. In this study, AFNI’s 3dAlleniate function was used for the regression process \cite{saad2009new}. The affine transformation parameters obtained were used to register resampled CT and MR images with the skull. To maximize information inside the brain area, CT images were windowed with a window length of 80 Hounsfield units (HU) and a window center of 40 HU. After registration (Fig. \ref{fig02}), CT and MR images were well-aligned spatially and temporally.
\subsection{MR-GAN}
\begin{figure}
  \centering
  \includegraphics[width=1\linewidth]{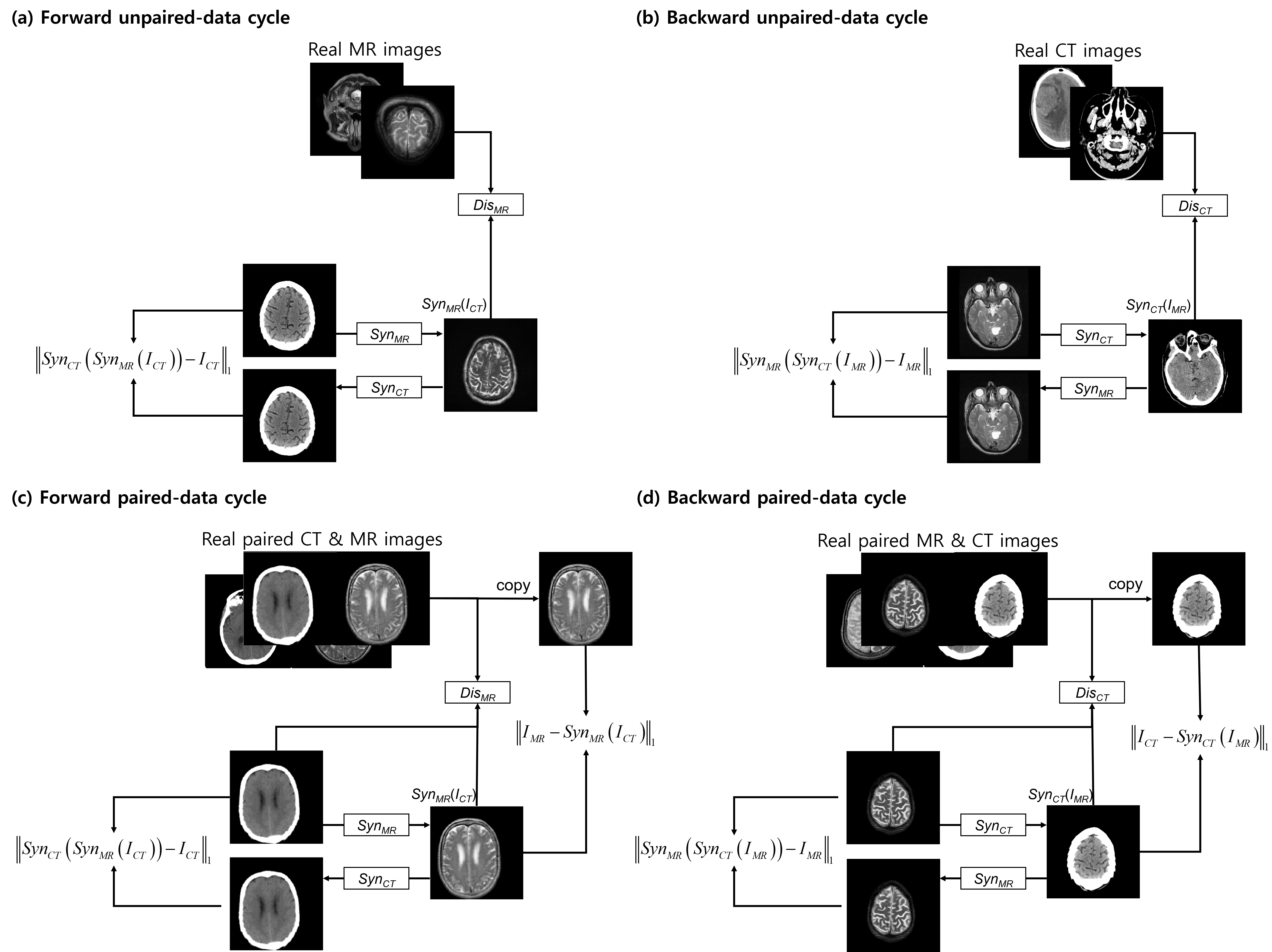}
  \caption{Dual cycle-consistent structure consists of (a) a forward unpaired-data cycle, (b) a backward unpaired-data cycle, (c) a forward paired-data cycle, and (d) a backward paired-data cycle. In the forward unpaired-data cycle, the input CT image is translated to an MR image by a synthesis network $Syn_{\!M\!R}$. The synthesized MR image is translated to a CT image that approximates the original CT image, and $Dis_{\!M\!R}$ is trained to distinguish between real and synthesized MR images. In the backward unpaired-data cycle, a CT image is instead synthesized from an input MR by the network $Syn_{CT}$, $Syn_{\!M\!R}$ reconstructs the MR from the synthesized CT image, and $Dis_{CT}$ is trained to distinguish between real and synthesized CT images. The forward paired-data and the backward paired-data cycle are the same as the above forward unpaired-data and the backward unpaired-data cycle. The difference is that $Dis_{\!M\!R}$ and $Dis_{CT}$ do not just discriminate between real and synthesized images, they learn to classify between real and synthesized pairs. In addition, the voxel-wise loss between the synthesized image and the reference image is included in the paired-data cycles.}
  \label{fig03}
\end{figure}

The proposed approach has a structure similar to CycleGAN \cite{zhu2017unpaired}, which contains a forward and a backward cycle. However, our model has a dual cycle-consistent term for paired and unpaired training data. The dual cycle-consistent term includes four cycles: forward unpaired-data, backward unpaired-data, forward paired-data, and backward paired-data cycles (Fig. \ref{fig03}).

The forward unpaired-data cycle contains three independent networks that each have a different goal. The network $Syn_{\!M\!R}$ attempts to translate a CT image $I_{CT}$ to a realistic MR image, so that the output cannot be distinguished from "real" MR images by the adversarially trained discriminator $Dis_{\!M\!R}$, which is trained to do as well as possible at discriminating the synthetic "fakes." In addition, to solve the well-known problem of mode collapse, the network $Syn_{CT}$ is trained to translate $Syn_{\!M\!R}\left(I_{CT}\right)$ back to the original CT domain. To improve training stability, the backward unpaired-data cycle is also enforced in, translating an MR image to a CT image, and it works with a logic opposite to the forward unpaired-data cycle. Unlike the unpaired-data cycle, the discriminators in the paired-data cycles do not just discriminate between real and synthesized images; they also observe a pair of CT and MR images to differentiate between real and synthesized pairs. In addition, the voxel-wise loss between the synthesized and the reference image is included in the paired-data cycles. The synthetic networks $Syn_{\!M\!R}$ and $Syn_{CT}$ in paired-data cycles work exactly the same as in the unpaired-data cycles.
\subsection{Objective}
Both networks in GAN were trained simultaneously with discriminators $Dis_{\!M\!R}$ and $Dis_{CT}$ estimating the probability that a sample came from real data rather than the synthesis networks, while the synthesis networks $Syn_{\!M\!R}$ and $Syn_{CT}$ were trained to translate realistic synthetic data that could not be distinguished from real data by the discriminators. We applied adversarial losses \cite{goodfellow2014generative} to the synthesis network $Syn_{\!M\!R}$: $I_{CT} \rightarrow I_{\!M\!R}$ and its discriminator $Dis_{\!M\!R}$, and express the objective as:
\begin{equation} \label{eq01}
\begin{split}
L_{G\!A\!N}\left(Syn_{\!M\!R},Dis_{\!M\!R},I_{CT},I_{\!M\!R}\right)&=\mathbb{E}_{I_{\!M\!R} \sim p_{d\!a\!t\!a}\left(I_{\!M\!R}\right)}\left[\log Dis_{\!M\!R}\left(I_{\!M\!R}\right)\right] \\
&+\mathbb{E}_{I_{CT} \sim p_{d\!a\!t\!a} \left(I_{CT}\right)}\left[\log \left(1 - Dis_{\!M\!R}\left(Syn_{\!M\!R}\left(I_{CT}\right)\right)\right)\right] \\
&+\mathbb{E}_{I_{CT}, I_{\!M\!R} \sim p_{d\!a\!t\!a}\left(I_{CT}, I_{\!M\!R} \right)}\left[\log Dis_{\!M\!R}\left(I_{CT},I_{\!M\!R}\right)\right] \\
&+\mathbb{E}_{I_{CT} \sim p_{d\!a\!t\!a}\left(I_{CT}\right)}\left[\log \left(1-Dis_{\!M\!R}\left(I_{CT},Syn_{\!M\!R}\left(I_{CT} \right)\right)\right)\right],
\end{split}
\end{equation}
where $Syn_{\!M\!R}$ tries to translate an $I_{CT}$ image to a $Syn_{\!M\!R}\left(I_{CT}\right)$ image that looks similar to an image from the MR image domain. For the first and the second term in Eq. (\ref{eq01}), the discriminator $Dis_{\!M\!R}$ aims to distinguish between synthesized $Syn_{\!M\!R}(I_{CT})$ and the real MR image $I_{\!M\!R}$ for unpaired data. For the paired data, the discriminator $Dis_{\!M\!R}$ also tries to discriminate between the real and synthesized pairs that provide $I_{CT}$ with the synthesized MR image as the third and fourth term in the Eq. (\ref{eq01}). The synthesis network $Syn_{\!M\!R}$ tries to minimize this objective against an adversarial $Dis_{\!M\!R}$ that tries to maximize it, i.e., $Syn_{\!M\!R}^*=\arg\min_{Syn_{\!M\!R}}\max_{Dis_{\!M\!R}}L_{G\!A\!N}\left(Syn_{\!M\!R},Dis_{\!M\!R},I_{CT},I_{\!M\!R}\right)$. For another synthesis network $Syn_{CT}$, $I_{\!M\!R} \rightarrow I_{CT}$ and discriminator $Dis_{CT}$ have a similar adversarial loss as well, i.e., $Syn_{\!C\!T}^*=\arg\min_{Syn_{\!C\!T}}\max_{Dis_{\!C\!T}}L_{G\!A\!N}\left(Syn_{\!C\!T},Dis_{\!C\!T},I_{\!M\!R},I_{CT}\right)$.

\newpage
To stabilize the training procedure, the negative log-likelihood objective in unpaired data was replaced by a least squares loss \cite{mao2017least} in our work. Hence, the discriminator $Dis_{\!M\!R}$ aims to apply the label 1 for real MR images, and the label 0 for synthesized MR images. However, we found that keeping the negative log-likelihood objective in the paired data generated higher quality results. Eq. (\ref{eq01}) then becomes:
\begin{equation} \label{eq02}
\begin{split}
L_{G\!A\!N}\left(Syn_{\!M\!R},Dis_{\!M\!R},I_{CT},I_{\!M\!R}\right)&=\mathbb{E}_{I_{\!M\!R} \sim p_{d\!a\!t\!a}\left(I_{\!M\!R}\right)}\left[\left(Dis_{\!M\!R}\left(I_{\!M\!R}\right)-1 \right)^2 \right] \\
&+\mathbb{E}_{I_{CT} \sim p_{d\!a\!t\!a} \left(I_{CT}\right)}\left[Dis_{\!M\!R}\left(Syn_{\!M\!R}\left(I_{CT}\right)\right)^2 \right] \\
&+\mathbb{E}_{I_{CT}, I_{\!M\!R} \sim p_{d\!a\!t\!a}\left(I_{CT}, I_{\!M\!R} \right)}\left[\log Dis_{\!M\!R}\left(I_{CT},I_{\!M\!R}\right)\right] \\
&+\mathbb{E}_{I_{CT} \sim p_{d\!a\!t\!a}\left(I_{CT}\right)}\left[\log \left(1-Dis_{\!M\!R}\left(I_{CT},Syn_{\!M\!R}\left(I_{CT}\right)\right)\right)\right]
\end{split}
\end{equation}
The dual cycle-consistent loss is enforced to further reduce the space of possible mapping functions for paired and unpaired training data. In the forward cycle, for each $I_{CT}$ from the CT domain, the image translation cycle should be able to bring $I_{CT}$ back to the original image, i.e., $I_{CT} \rightarrow Syn_{\!M\!R}\left(I_{CT}\right) \rightarrow Syn_{CT}\left(Syn_{\!M\!R}\left(I_{CT}\right)\right) \approx I_{CT}$. Similarly, for each $I_{\!M\!R}$ from the MR domain, $Syn_{CT}$ and $Syn_{\!M\!R}$ should also satisfy a backward cycle consistency: $I_{\!M\!R} \rightarrow Syn_{CT}\left(I_{\!M\!R}\right) \rightarrow Syn_{\!M\!R}\left(SynCT\left(I_{\!M\!R}\right)\right) \approx I_{\!M\!R}$. The dual cycle-consistency loss is expressed as:
\begin{equation}\label{eq03}
\begin{split}
L_{dual-cyc}\left(Syn_{\!M\!R},Syn_{CT}\right)=&\mathbb{E}_{I_{CT} \sim p_{d\!a\!t\!a}\left(I_{CT}\right)}\left[
\|Syn_{CT}\left(Syn_{\!M\!R}\left(I_{CT}\right)\right)-I_{CT}\|_1\right] \\
&+\mathbb{E}_{I_{\!M\!R} \sim p_{d\!a\!t\!a} \left(I_{\!M\!R}\right)}\left[\|Syn_{\!M\!R}\left(Syn_{CT}\left(I_{\!M\!R}\right)\right)-I_{\!M\!R}\|_1\right] \\
&+\mathbb{E}_{I_{CT}, I_{\!M\!R} \sim p_{d\!a\!t\!a}\left(I_{CT}, I_{\!M\!R} \right)}\left[\|Syn_{CT}\left(Syn_{\!M\!R}\left(I_{CT}\right)\right)-I_{CT}\|_1\right] \\
&+\mathbb{E}_{I_{\!M\!R},I_{CT} \sim p_{d\!a\!t\!a}\left(I_{\!M\!R},I_{CT}\right)}\left[\|Syn_{\!M\!R}\left(Syn_{CT}\left(I_{\!M\!R}\right)\right)-I_{\!M\!R}\|_1\right]
\end{split}
\end{equation}
Previous approaches \cite{pathak2016context} have found it beneficial to combine the adversarial loss with a more traditional loss, such as L1 distance. For the paired data {$I_{CT}$, $I_{\!M\!R}$}, the synthesis network $Syn_{\!M\!R}$ is tasked to not only generate realistic MR images, but also to be near the reference $I_{\!M\!R}$ of the input $I_{\!C\!T}$. Though we don’t need a synthesis network $Syn_{CT}$ as a final product, adding the same constraint to the $Syn_{CT}$ enables a higher quality of synthesized MR images. The L1 loss term for the $Syn_{\!M\!R}$ and $Syn_{CT}$ are:
\begin{equation}\label{eq04}
\begin{split}
L_{L1}\left(Syn_{\!M\!R},Syn_{CT}\right)=&\mathbb{E}_{I_{CT},I_{\!M\!R}\sim p_{d\!a\!t\!a}\left(I_{CT},I_{\!M\!R}\right)}\left[\|I_{\!M\!R}-Syn_{\!M\!R} \left(I_{CT}\right)\|_1\right]\\
&+\mathbb{E}_{I_{\!M\!R},I_{CT}\sim p_{d\!a\!t\!a}\left(I_{\!M\!R},I_{CT}\right)}\left[\|I_{CT}-Syn_{CT} \left(I_{\!M\!R}\right)\|_1\right]
\end{split}
\end{equation}
The overall objective is:
\begin{equation}\label{eq05}
\begin{split}
L\left(Syn_{\!M\!R},Syn_{CT},Dis_{\!M\!R},Dis_{CT}\right)=&L_{\!G\!A\!N}\left(Syn_{\!M\!R},Dis_{\!M\!R},I_{CT},I_{\!M\!R}\right)\\
&+L_{\!G\!A\!N}\left(Syn_{CT},Dis_{CT},I_{\!M\!R},I_{CT}\right)\\
&+\lambda L_{dual-cyc}\left(Syn_{\!M\!R},Syn_{CT}\right)\\
&+\gamma L_{L1}\left(Syn_{\!M\!R},Syn_{CT}\right)
\end{split}
\end{equation}
where $\lambda$ and $\gamma$ control the relative importance of adversarial loss, dual cycle-consistent loss, and voxel-wise loss. We aim to solve the Eq. (\ref{eq06}): 
\begin{equation}\label{eq06}
Syn_{\!M\!R}^* = \arg \min_{Syn_{\!M\!R},Syn_{CT}} \max_{Dis_{\!M\!R,Dis_{CT}}} L\left(Syn_{\!M\!R},Syn_{CT},Dis_{\!M\!R},Dis_{CT}\right).
\end{equation}
The MR-GAN procedure is described in Algorithm \ref{alg:MRGAN}.

\begin{algorithm}[H] 
\caption{MR-GAN, proposed algorithm. All experiments in the paper used the default values $m=1$, $n_{inter}=1$.} 
\label{alg:MRGAN}
\begin{algorithmic}[1] 
\Require $\alpha$, the learning rate. $m$, the batch size. $n_{inter}$, the number of iterations of the unpaired/paired data.
\For{number of training iterations}
  \For{$n_{iter}$ steps}
    \State Sample $\left\{I_{CT}^{\left(i\right)}\right\}_{i=1}^{m} \sim \mathcal{P}_{data} \left(I_{CT}\right)$ a batch from the unpaired CT data.
    \State Sample $\left\{I_{\!M\!R}^{\left(i\right)}\right\}_{i=1}^{m} \sim \mathcal{P}_{data}\left(I_{\!M\!R}\right)$ a batch from the unpaired MR data.
    \State Update the discriminator, $Dis_{\!M\!R}$, by ascending its stochastic gradient:
    $$\bigtriangledown_{\theta_d^{\!M\!R}}\frac{1}{m} \sum_{i=1}^{m}  \left[\left(Dis_{\!M\!R}\left(I_{\!M\!R}^{(i)}\right)-1 \right)^2 + Dis_{\!M\!R}\left(Syn_{\!M\!R}\left(I_{CT}^{(i)}\right)\right)^2\right].$$
    \State Update the generator, $Syn_{\!M\!R}$, by descending its stochastic gradient:
    $$\bigtriangledown_{\theta_g^{\!M\!R}}\frac{1}{m} \sum_{i=1}^{m} \left[Dis_{\!M\!R}\left(Syn_{\!M\!R}\left(I_{CT}^{(i)}\right)\right)^2 + \big\|Syn_{CT}\left(Syn_{\!M\!R}\left(I_{CT}^{(i)}\right)\right)-I_{CT}^{(i)}\big\|_1 \right].$$
    \State Update the discriminator, $Dis_{CT}$, by ascending its stochastic gradient:
    $$\bigtriangledown_{\theta_d^{CT}}\frac{1}{m} \sum_{i=1}^{m}  \left[\left(Dis_{CT}\left(I_{CT}^{(i)}\right)-1 \right)^2 + Dis_{CT}\left(Syn_{CT}\left(I_{\!M\!R}^{(i)}\right)\right)^2\right].$$
    \State Update the generator, $Syn_{CT}$, by descending its stochastic gradient:
    $$\bigtriangledown_{\theta_g^{CT}}\frac{1}{m} \sum_{i=1}^{m} \left[Dis_{CT}\left(Syn_{CT}\left(I_{\!M\!R}^{(i)}\right)\right)^2 + \big\|Syn_{\!M\!R}\left(Syn_{CT}\left(I_{\!M\!R}^{(i)}\right)\right)-I_{\!M\!R}^{(i)}\big\|_1 \right].$$
  \EndFor
\algstore{myalg}
\end{algorithmic}
\end{algorithm}

\begin{algorithm}[H]
\begin{algorithmic}[1]
\algrestore{myalg}
  \For{$n_{iter}$ steps}
    \State Sample $\left\{I_{CT}^{\left(i\right)}, \ I_{\!M\!R}^{\left(i\right)} \right\}_{i=1}^{m} \sim \mathcal{P}_{data} \left(I_{CT}, I_{\!M\!R}\right)$ a batch from the paired data.
    \State Update the discriminator, $Dis_{\!M\!R}$, by ascending its stochastic gradient:
    $$\bigtriangledown_{\theta_d^{\!M\!R}}\frac{1}{m} \sum_{i=1}^{m} \left[\log Dis_{\!M\!R}\left(I_{CT}^{(i)},I_{\!M\!R}^{(i)} \right) +  \log \left(1-Dis_{\!M\!R}\left(I_{CT}^{(i)},Syn_{\!M\!R}\left(I_{CT}^{(i)}\right)\right)\right) \right]$$
    \State Update the generator, $Syn_{\!M\!R}$, by descending its stochastic gradient:
\begin{equation*}
\begin{split}
\bigtriangledown_{\theta_g^{\!M\!R}}\frac{1}{m} \sum_{i=1}^{m} &\log \left(1-Dis_{\!M\!R}\left(I_{CT}^{(i)},Syn_{\!M\!R}\left(I_{CT}^{(i)}\right)\right)\right) + \big \|Syn_{CT}\left(Syn_{\!M\!R}\left(I_{CT}^{(i)}\right)\right)-I_{CT}^{(i)} \big \|_1 \\
&+ \big \|I_{\!M\!R}^{(i)}-Syn_{\!M\!R} \left(I_{CT}^{(i)}\right) \big \|_1
\end{split}
\end{equation*}
    \State Update the discriminator, $Dis_{CT}$, by ascending its stochastic gradient:
    $$\bigtriangledown_{\theta_d^{CT}}\frac{1}{m} \sum_{i=1}^{m} \left[\log Dis_{CT}\left(I_{\!M\!R}^{(i)},I_{CT}^{(i)} \right) +  \log \left(1-Dis_{CT}\left(I_{\!M\!R}^{(i)},Syn_{CT}\left(I_{\!M\!R}^{(i)}\right)\right)\right) \right]$$
    \State Update the generator, $Syn_{CT}$, by descending its stochastic gradient:
    \begin{equation*}
    \begin{split}
    \bigtriangledown_{\theta_g^{CT}}\frac{1}{m} \sum_{i=1}^{m} &\log \left(1-Dis_{CT}\left(I_{\!M\!R}^{(i)},Syn_{CT}\left(I_{\!M\!R}^{(i)}\right)\right)\right) + \big \|Syn_{\!M\!R}\left(Syn_{CT}\left(I_{\!M\!R}^{(i)}\right)\right)-I_{\!M\!R}^{(i)} \big \|_1 \\
     &+ \big \|I_{CT}^{(i)}-Syn_{CT} \left(I_{\!M\!R}^{(i)}\right) \big \|_1
    \end{split}
    \end{equation*}
  \EndFor
\EndFor \\
\Return $\text{result}$
\end{algorithmic}
\end{algorithm}

\begin{figure}
  \centering
  \includegraphics[width=0.8\linewidth]{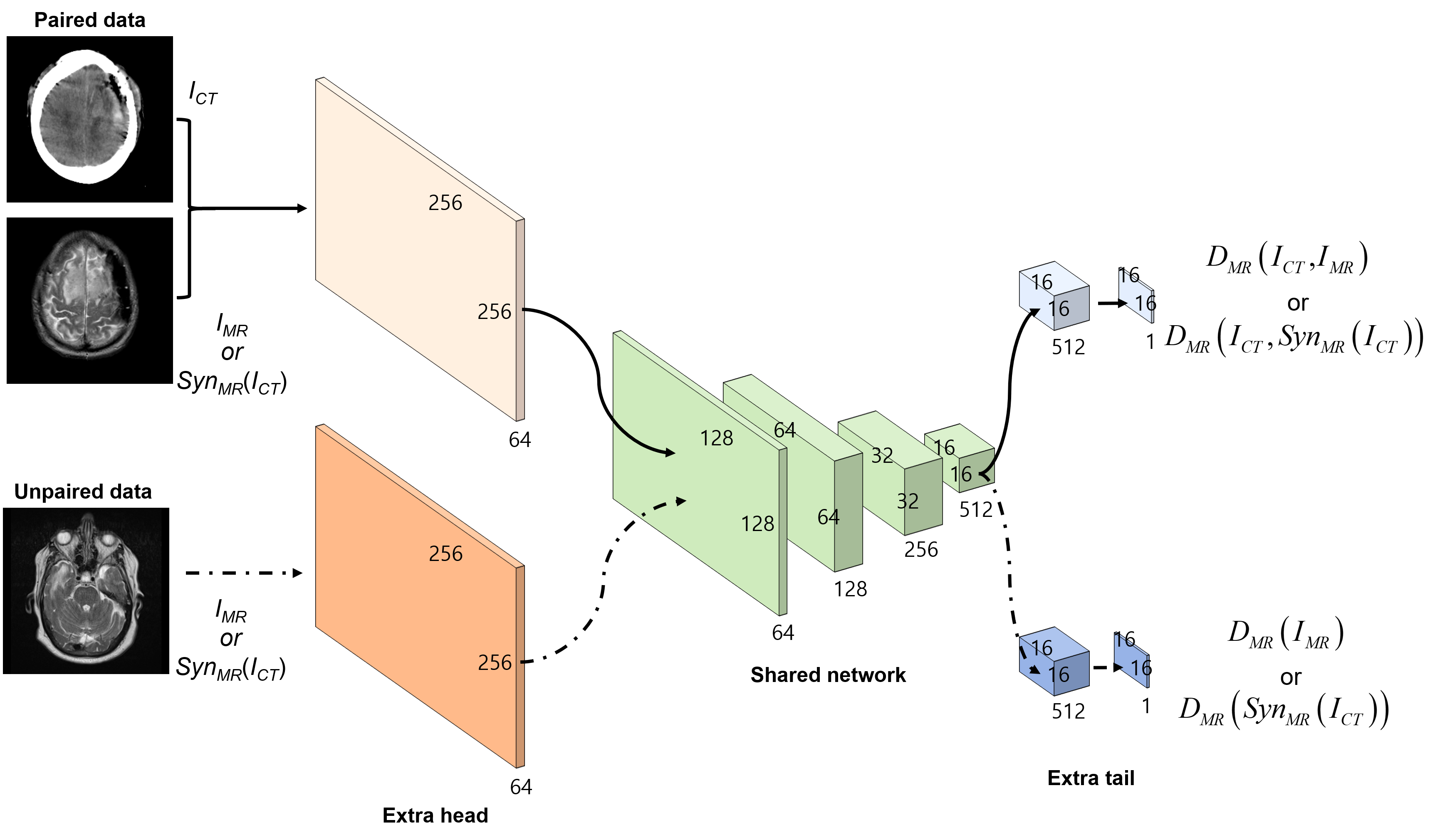}
  \caption{Flow diagram of the discriminator $Dis_{\!M\!R}$ in the synthetic system. $Dis_{\!M\!R}$ has extra head and extra tail convolutional layers for the different input and loss functions of the paired and unpaired data. Discriminator $Dis_{CT}$ has the same architecture as the $Dis_{\!M\!R}$.}
  \label{fig04}
\end{figure} 
\subsection{Implementation}
For the architecture of synthesis networks $Syn_{\!M\!R}$ and $Syn_{CT}$, we utilized the archiecture from Johnson et al. \cite{johnson2016perceptual}, which was a 2D fully-convolutional network with one convolutional layer, followed by two strided convolutional layers, nine residual blocks \cite{he2016deep}, two fractionally-strided convolutional layers, and one last convolutional layer. Instance normalization \cite{ulyanovinstance} and ReLU followed each convolution except at the last convolutional layer. The synthesis network takes a $256\times256$ input and generates an output image of the same size.

For the discriminators $Dis_{\!M\!R}$ and $Dis_{CT}$, we adapted PatchGANs \cite{isola2017image}, which tries to classify each $N \times N$ patch in an image as real or fake. This way, the discriminators could better focus on high-frequency information in local image patches. Networks $Dis_{\!M\!R}$ and $Dis_{CT}$ used the same architecture, which had one convolution as an extra head for different input data, four strided convolutions as a shared network, and two convolutions as an extra tail for different tasks. Except for the first and last convolution, each convolutional layer was followed by instance normalization \cite{ulyanovinstance} and leaky ReLu \cite{xu2015empirical} (Fig. \ref{fig04}).

To optimize our networks, we used minibatch SGD and applied the Adam optimizer \cite{kingma2014adam} with a batch size of 1. The learning rate started at $2e^{-4}$ for the first $1e^5$ iterations, and decayed linearly to zero over the next $2e^5$ iterations. For all experiments, we set $\lambda = 10$ and $\gamma = 100$ in Eq. (\ref{eq05}) empirically. At inference time, we ran the synthesis network $Syn_{\!M\!R}$ only to give a CT image.

\section{Results}
\subsection{Data preprocessing}
Among the data of $202$ patients, all of the unpaired data were used as training data. The paired data were separated into a training set with the data of $10$ patients, and a separate test set containing CT images, and corresponding reference MR images from $10$ patients. Each CT or MR volume involved more than $35$ 2D axial image slices. These were resampled to $256 \times 256$ in $256$-grayscale and uniformly distributed by HU for CT and MR data.

For training, we augmented the training data with random online transforms:
\begin{itemize}
\setlength\itemsep{0em}
\item {\it Flip}: Batch data were horizontally flipped with $0.5$ probability.
\item {\it Translation}: Batch data were randomly cropped to size $256 \times 256$ from padded $286 \times 286$.
\item {\it Rotation}: Batch data were rotated by $r\in\left[-5, 5\right]$ degrees.
\end{itemize}
The paired CT and MR images were augmented using the same factor. However, in the unpaired data, CT and MR images were augmented independently. The proposed approach training took about $72$ hours for $3e^5$ iterations using a single GeForce GTX 1080Ti GPU. At inference time, the system required $35$ ms to synthesize a single-slice CT image to MR image.

\subsection{Evaluation metrics}
Real and synthesized MR images were compared using the mean absolute error (MAE)
\begin{equation}\label{eq09}
M\!A\!E=\frac{1}{N} \sum\limits_{i=0}^{N-1}\|I_{\!M\!R}\left(i\right)-Syn_{\!M\!R}\left(I_{CT}\left(i\right)\right)\|,
\end{equation}
where $i$ is the index of the 2D axial image slice in aligned voxels, and $N$ is the number of slices in the reference MR images. MAE measures the average distance between each pixel of the synthesized and the reference MR image. In addition, the synthesized MR images were evaluated using the peak-signal-to-noise-ratio (PSNR) as proposed in \cite{
nie2017medical, bi2017synthesis, wolterink2017deep}:
\begin{equation}\label{eq10}
P\!S\!N\!R=10 \cdot \log_{10}\left(\frac{M\!A\!X^2}{M\!S\!E}\right),
\end{equation}
\begin{equation}\label{eq11}
M\!S\!E=\frac{1}{N} \sum\limits_{i=0}^{N-1}\left(I_{\!M\!R}\left(i\right)-Syn_{\!M\!R}\left(I_{CT}\left(i\right)\right)\right)^2,
\end{equation}
where $M\!A\!X=255$. PSNR measures the ratio between the maximum possible intensity value and the mean square error (MSE) of the synthesized and reference MR images.
\begin{figure}
  \centering
  \includegraphics[width=1.\linewidth]{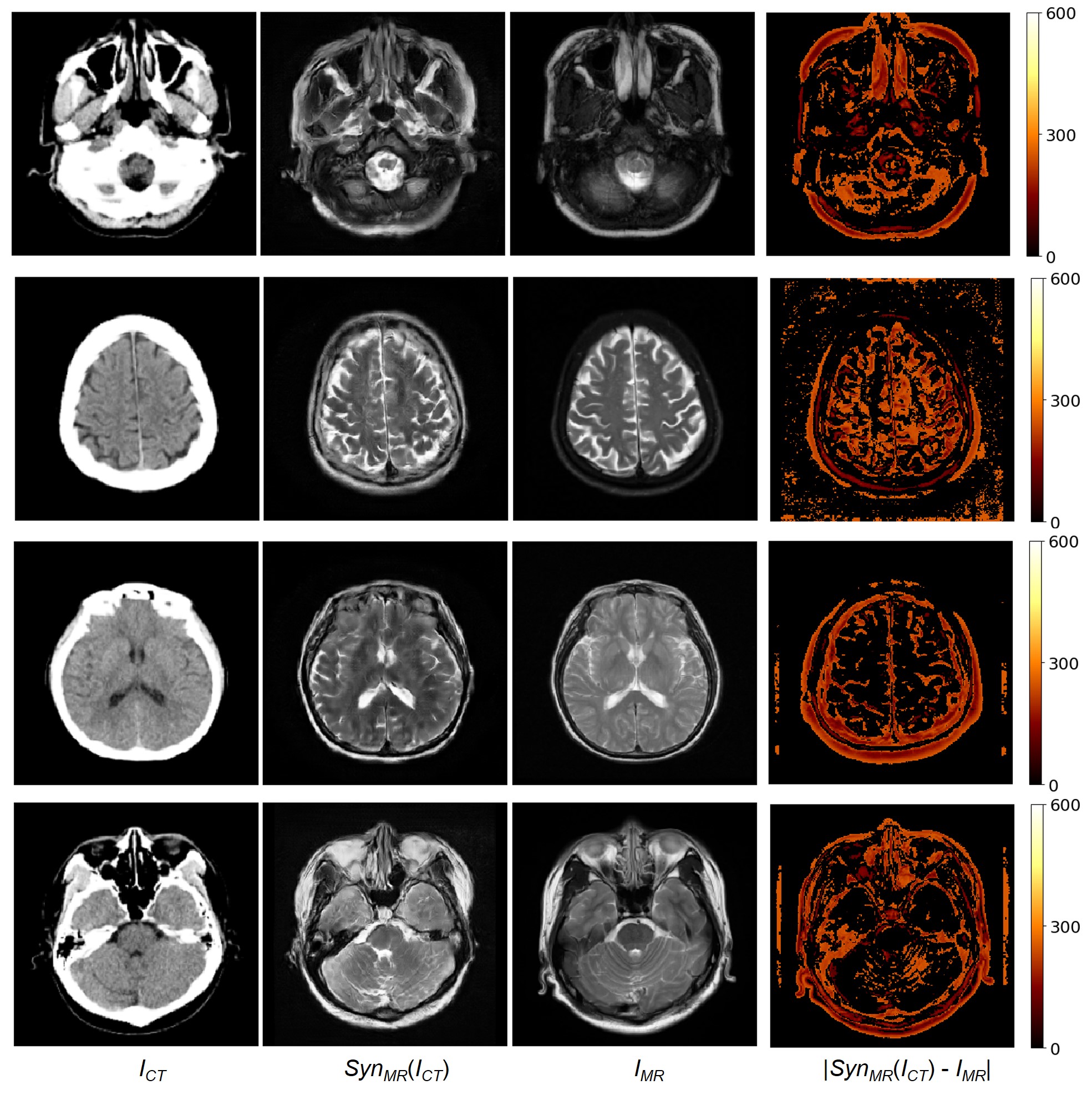}
  \caption{{\it From left to right} Input CT, synthesized MR, reference MR, and absolute error between real and synthesized MR images.}
  \label{fig05}
\end{figure}
\subsection{Analysis of MR synthesis using paired and unpaired data}
We first compared synthesized MR images with reference MR images that had been carefully registered to become paired data with CT images. For brevity, we refer to our method as MR-GAN. Fig. \ref{fig05} shows four examples of an input CT image, synthesized MR image obtained by MR-GAN, reference MR image, and absolute difference maps between the synthesized and reference MR images. The MR-GAN learned to differentiate between different anatomical structures with similar pixel intensity in CT images, such as bones, gyri, and soft brain tissues. The largest differences are in the area of bony structures, and the smallest differences are found in the soft brain tissues. This may be partly due to the misalignment between the CT and reference MR images, and because the CT image provides more detail about bony structures to complement the shortcoming of the synthesized MR, which is focused on soft brain tissues.

Table \ref{tab01} shows a quantitative evaluation using MAE and PSNR to compare different methods in the test set. We compare the proposed method with independent training using paired and unpaired data. To train the paired data system, a synthesis network with the same architecture $Syn_{\!M\!R}$ and a discriminator network with the same architecture $Dis_{\!M\!R}$ are trained using a combination of adversarial loss and voxel-wise loss as in the pix2pix framework \cite{isola2017image}. To train the unpaired data system \cite{wolterink2017deep}, the cycle-consistent structure of the CycleGAN \cite{zhu2017unpaired} model is used, which is the same as our approach for the forward and backward unpaired-data cycle, shown in Fig. \ref{fig03}. To ensure a fair comparison, we implemented all the baselines using the same architecture and implementation details as our method.

\begin{table}
  \caption{MAE and PSNR evaluations between synthesized and real MR images when training with paired, unpaired, and paired with unpaired data (Ours).}
  \label{tab01}
  \centering
  \begin{tabular}{L{2.2cm}C{2.2cm}C{2.2cm}C{2.2cm}C{2.2cm}C{2.2cm}C{2.2cm}}
    \hline \hline
    \multicolumn{1}{c}{} &
    \multicolumn{3}{c}{MAE} & 
    \multicolumn{3}{c}{PSNR} \\ [-0.5em]
    {} &Paired &Unpaired &Ours &Paired &Unpaired &Ours \\
    \midrule
    \hline
    Patient01 &24.20 &27.71 &22.76 &62.82 &62.45 &64.65 \\ [-0.5em]
    Patient02 &17.82 &24.12 &18.27 &64.91 &63.05 &65.93 \\ [-0.5em]
    Patient03 &22.01 &22.45 &22.27 &63.59 &63.83 &63.55 \\ [-0.5em]
    Patient04 &18.23 &23.64 &16.75 &65.28 &63.44 &65.76 \\ [-0.5em]
    Patient05 &18.26 &22.82 &17.68 &64.92 &64.04 &65.97 \\ [-0.5em]
    Patient06 &20.52 &20.41 &17.57 &64.87 &64.78 &65.92 \\  [-0.5em]
    Patient07 &20.63 &18.72 &16.55 &64.55 &64.14 &66.28 \\ [-0.5em]
    Patient08 &19.42 &22.77 &18.30 &64.10 &63.22 &65.82 \\ [-0.5em]
    Patient09 &19.12 &16.98 &18.57 &64.93 &66.19 &65.43 \\ [-0.5em]
    Patient10 &23.23 &29.76 &24.91 &63.81 &62.60 &64.17 \\ [-0.5em]
    Avg$\pm$sd &20.34$\pm$2.20 &22.94$\pm$3.62 &19.36$\pm$2.73 &64.28$\pm$0.81 &63.77$\pm$1.06 &65.35$\pm$0.86 \\
    \hline \hline
  \end{tabular}
\end{table}
\begin{table}
  \caption{Network architecture} \label{ta:tab02}
  \centering
  \begin{tabular}{C{3cm}C{2.4cm}C{2.4cm}C{2.4cm}C{2.4cm}C{2.4cm}C{2.4cm}}
    \hline \hline
    Discriminator &D1 &D2 &D3 &D4 &D5 \\
    \hline
    Extra head &(64) &(64, 64) &(64, 64) &(64) &(64) \\
    Shared network &\thead{(64, 128, \\[-0.5em] 256, 512)} &\thead{(64, 128, \\[-0.5em] 256, 512)} &\thead{(64, 128, \\[-0.5em] 256, 512)} &\thead{(128, 256, \\[-0.5em] 512)} &\thead{(128, 256, \\[-0.5em] 512)} \\
    Extra tail &(512, 1) &\thead{(512, \\[-0.5em] 512, 1)} &\thead{(512, 512,\\[-0.5em] 512, 1)} &(512, 1) &(1) \\
    \hline \hline
  \end{tabular}
\end{table}
\begin{table}
  \caption{Comparison of the MAE and PSNR for different discriminator networks and leaset squares loss. The leading scores are displayed in bold font.} \label{ta:tab03}
  \centering
  \begin{tabular}{C{5cm}C{3.5cm}C{3.5cm}C{3.5cm}}
  \hline \hline
  \multirow{2}{*}{Model} &\multicolumn{3}{c}{Leaset squares loss} \\ [-0.5em]
  &D1 &D2 &D3 \\
  \hline
  MAE (Avg$\pm$sd) &\textbf{21.07$\pm$2.98} &42.95$\pm$3.03 &37.25$\pm$2.58 \\ [-0.5em]
  PSNR (Avg$\pm$sd) &\textbf{65.25$\pm$0.81} &61.31$\pm$0.64 &62.73$\pm$0.77 \\
  \hline \hline
  \end{tabular}
\end{table}
\begin{table}
  \caption{Comparison of the MAE and PSNR for different discriminator networks and negative log-likelihood. The leading scores are displayed in bold font.} \label{ta:tab04}
  \centering
  \small
  \begin{tabular}{C{3cm}C{2.5cm}C{2.5cm}C{2.5cm}C{2.5cm}C{2.5cm}}
  \hline \hline
  \multirow{2}{*}{Model} &\multicolumn{5}{c}{Negative log-likelihood} \\ [-0.5em]
  &D1 &D2 &D3 &D4 &D5 \\
  \hline
  MAE (Avg$\pm$sd) &\textbf{19.36$\pm$2.73} &49.70$\pm$3.10 &59.06$\pm$3.27 &\textbf{19.35$\pm$2.56} &20.57$\pm$2.82 \\ [-0.5em]
  PSNR (Avg$\pm$sd) &\textbf{65.35$\pm$0.86} &60.34$\pm$0.60 &59.23$\pm$0.46 &\textbf{65.24$\pm$0.77} &65.16$\pm$0.85 \\
  \hline \hline
  \end{tabular}
\end{table}
\begin{figure}
  \centering
  \includegraphics[width=1\linewidth]{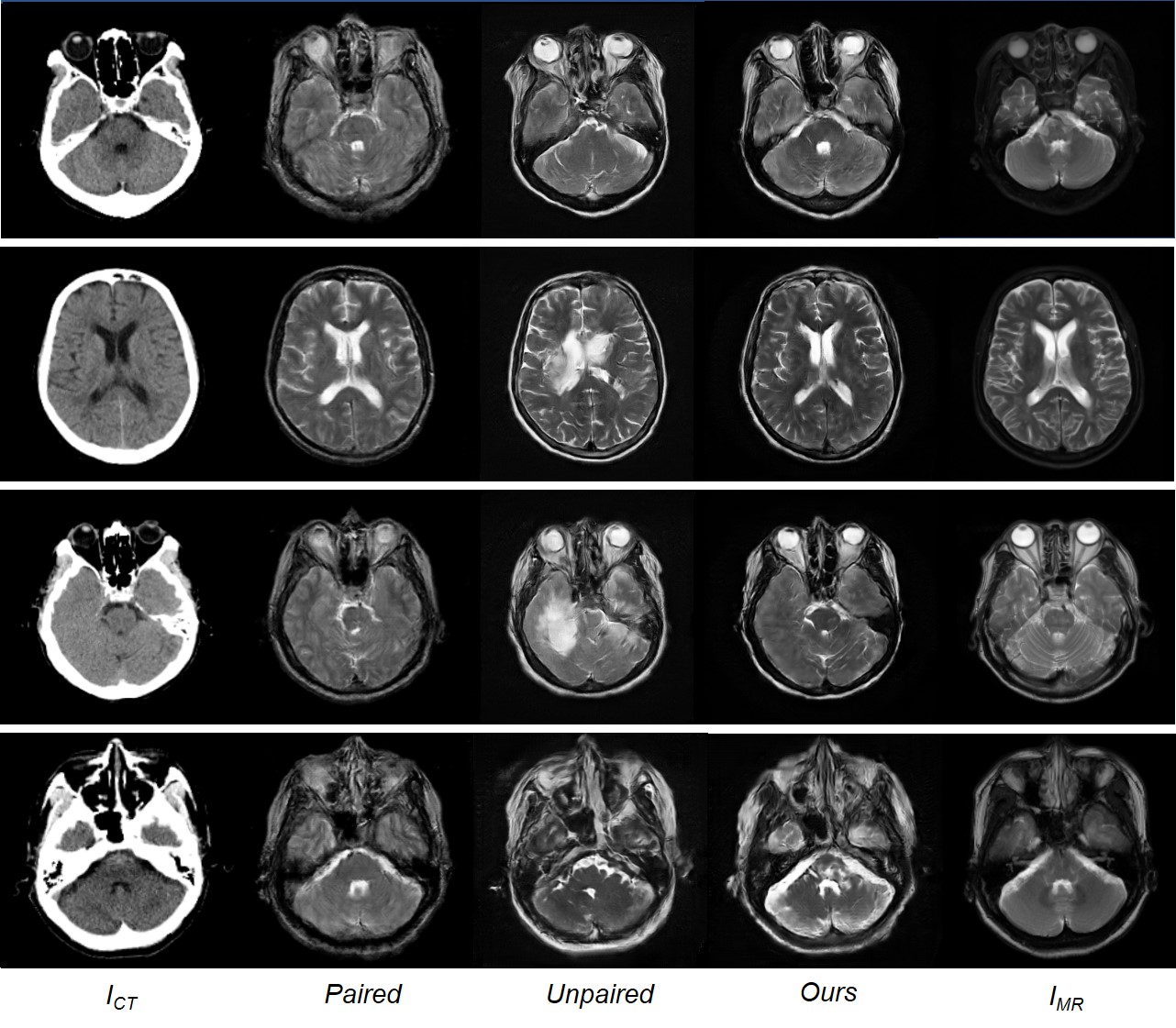}
  \caption{{\it From left to right} Input CT image, synthesized MR image with paired training, synthesized MR image with unpaired training, synthesized MR images with paired and unpaired training (ours), and reference MR images.}
  \label{fig06}
\end{figure}
\begin{figure}
  \centering
  \includegraphics[width=1\linewidth]{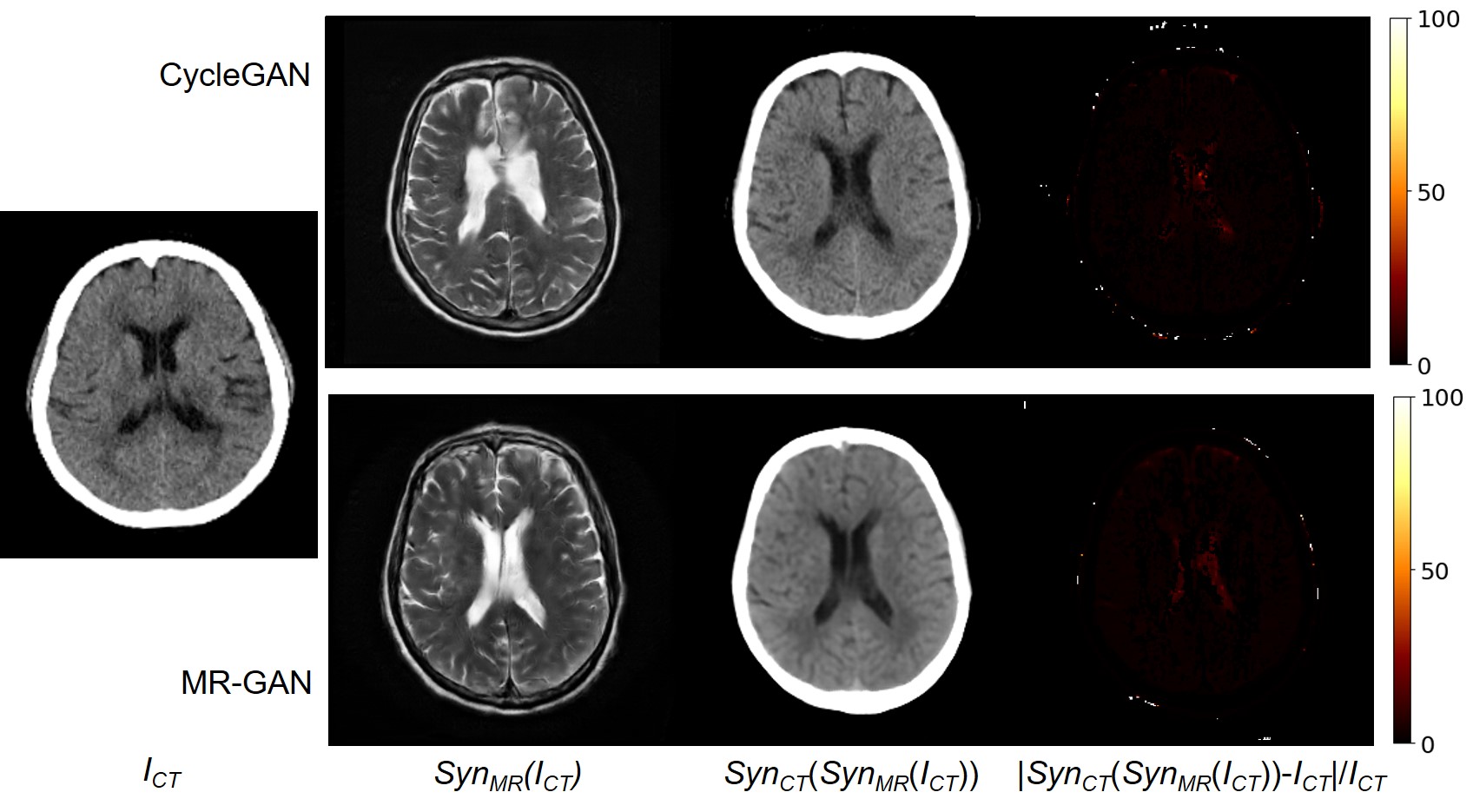}
  \caption{{\it From left to right} Input CT image, synthesized MR image, reconstructed CT image, and relative difference error between the input and reconstructed CT image.}
  \label{fig07}
\end{figure}

\newpage
Although having trained with limited paired data, the model using paired training data outperformed the CycleGAN model on unpaired data in our experiments. Table \ref{tab01} indicates that our approach to training with paired and unpaired data together had the best performance across all measurements, with the lowest MAE and highest PSNR compared to the conventional paired and unpaired training. Fig. \ref{fig06} shows a qualitative comparison between paired training, unpaired training, and our approach. The results of training with paired data seemed good, but generated blurry outputs. The images obtained with unpaired training were realistic, but lost anatomical information in ares of soft brain tissue, and contained artifacts in areas with bony structures. Although our method learns translation using paired and unpaired data, the quality of our results closely approximates the reference MR images, and for some details our results are much clearer than the reference MR images.

We present comparison results for several discriminator models. As mentioned in Fig. \ref{fig04}, the discriminator is consists of the extra head, shared network, and extra tail. Different discriminator models are presented in Table \ref{ta:tab02}, use standard $4 \times 4$ padded convolution with stride 1. The comparison of the MAE and PSNR for different discriminator networks and objective functions are given in Table \ref{ta:tab03} and Table \ref{ta:tab04}. The results clearly indicate that the discriminators with the negative log-likelihood better than least squares loss \cite{mao2017least} in terms of MAE and PSNR. We observed that the performance was positively correlated with the number of convolution layers in the extra head and extra tail of the discriminator. With little convolution layers in the two networks, the discriminator prevents overfitting in the paired and unpaired learning. We also noted that the performance was uncorrelated to the number of convolution layers in the shared network.

During the training of the MR-GAN, dual cycle-consistency is explicitly imposed in a bi-direction way. Hence, an input CT image translated to an MR image by the model should be successfully translated back to the original CT domain. Fig. \ref{fig07} shows an input CT, corresponding synthesized MR images from the CycleGAN and MR-GAN, their reconstructed CT images, and their relative difference maps. We observed that the reconstructed CT images were very close to the input images. Relative differences are distributed at the contour of the bone, and the reconstructed CT image from MR-GAN is more smoothed than the CycleGAN model because of the correct SynMR(ICT), which is like a latent vector in an auto-encoder \cite{hinton2006reducing}.

\section{Discussion}
This paper has shown that a synthetic system can be trained using paired and unpaired data to synthesize an MR images from a CT image. Unlike other methods, the proposed approach utilizes the adversarial loss from a discriminator network, dual cycle-consistent loss using paired and unpaired training data, and voxel-wise loss based on paired data to synthesize realistically-looking MR images. Quantitative evaluation results in Table \ref{tab01} show that the average correspondence between synthesized and reference MR images in our approach is much better than in other methods; synthesized images are closer to the reference, and achieve the lowest MAE of $19.36 \pm 2.73$ and the highest PSNR of $65.35 \pm 0.86$. Slight misalignments between CT images and reference MR images may have a large effect on quantitative evaluation. Although a quantitative measurement may be the gold standard for assessing the performance of a method, we found that numerical differences in the quantitative evaluation do not indicate the qualitative difference correctly. In future work, we will evaluate the accuracy of synthesized MR images based on perceptual studies with medical experts.

A synthetic system using a CycleGAN model \cite{zhu2017unpaired} and trained with unpaired data generated realistic results. However, the results had poor anatomical definitions compared with corresponding CT images, as exemplified in Fig. \ref{eq06}. We found that even though it was trained with limited paired data, the pix2pix model \cite{isola2017image} outperformed the CycleGAN model on unpaired data in our experiments. The limitation of paired training is blurry output due to the voxel-wise loss. Qualitative analysis showed that MR images obtained by the MR-GAN look more realistic and contain less blurring than other methods. This could be due to the dual cycle-consistent and voxel-wise loss for paired data.

The experimental results have implications for accurate CT-based radiotherapy treatment for patients who are contraindicated to undergo an MR scan because of cardiac pacemakers or metal implants, and patients who live in areas with poor medical services. Our synthetic system can be trained using any kind of data: paired, unpaired, or both. Using paired and unpaired data together obtain higher quality synthesized images than using one kind of data alone.

\newpage
\section{Conclusion}
We propose a system for synthesizing MR images from CT images. Our approach uses paired and unpaired data to solve the context-misalignment problem of unpaired training, and alleviate the rigid registration task and blurred results of paired training. Unpaired data is plentifully available, and together with limited paired data, could be used for effective synthesis in many cases. Our results on the test set demonstrate that MR-GAN was much closer to the reference MR images when compared with other methods. The preliminary results indicated that the synthetic system is able to efficiently translate structures within complicated 2D brain slices, such as soft brain vessels, gyri, and bones. In future work, we will investigate the 3D information of anatomical structures as presented in CT and MR brain sequences to further improve performance based on paired and unpaired data. We suggest that our approach can potentially increase the quality of synthesized images for a synthetic system that depends on supervised and unsupervised settings, and can also be extended to support other applications, such as MR-CT and CT-PET synthesis.


%
%

%



\end{document}